\documentclass[sigconf,review=false]{acmart}

\AtBeginDocument{%
  \providecommand\BibTeX{{%
    \normalfont B\kern-0.5em{\scshape i\kern-0.25em b}\kern-0.8em\TeX}}}

\copyrightyear{2023}
\acmYear{2023}
\setcopyright{rightsretained}

\acmDOI{}
\acmISBN{}

\usepackage{eurosym}
\usepackage{multirow}
\usepackage{bm}
\usepackage{bbm}
\usepackage{xcolor}

\begin{document}


\title{Incremental Profit per Conversion: a Response Transformation for Uplift Modeling in E-Commerce Promotions}
\author{Hugo Manuel Proen\c{c}a}
\email{hugo.proenca@booking.com}
\orcid{0000-0001-7315-5925}
\affiliation{%
  \institution{Booking.com}
  \city{Amsterdam}
  \country{Netherlands}
}

\author{Felipe Moraes}
\email{felipe.moraes@booking.com}
\affiliation{%
  \institution{Booking.com}
  \city{Amsterdam}
  \country{Netherlands}
}


\begin{abstract}
Promotions play a crucial role in e-commerce platforms, and various cost structures are employed to drive user engagement. This paper focuses on promotions with response-dependent costs, where expenses are incurred only when a purchase is made. Such promotions include discounts and coupons. While existing uplift model approaches aim to address this challenge, these approaches often necessitate training multiple models, like meta-learners, or encounter complications when estimating profit due to zero-inflated values stemming from non-converted individuals with zero cost and profit.

To address these challenges, we introduce Incremental Profit per Conversion (IPC), a novel uplift measure of promotional campaigns' efficiency in unit economics. Through a proposed response transformation, we demonstrate that IPC requires only converted data, its propensity, and a single model to be estimated. As a result, IPC resolves the issues mentioned above while mitigating the noise typically associated with the class imbalance in conversion datasets and biases arising from the many-to-one mapping between search and purchase data. Lastly, we validate the efficacy of our approach by presenting results obtained from a synthetic simulation of a discount coupon campaign.

\end{abstract}







\maketitle
\settopmatter{printacmref=false} 
\renewcommand\footnotetextcopyrightpermission[1]{} 
\pagestyle{plain} 
\section{Introduction}


Promotions are a common way for e-commerce platforms to invest their marketing budget in customer acquisition~\cite{rossler2022sharing} and churn prevention~\cite{lemmens2020managing, devriendt2021you}. E-commerce industry has traditionally estimated the causal effects of product changes through controlled experiments, such as A/B tests, where users are randomly split between two groups, with one receiving the original version A and the other the new version B~\cite{kaufman2017democratizing}. This methodology can also be applied to study the effectiveness of promotion campaigns, where version A is again the control group, but version B receives the promotion. Thus, if the promotion's Average Treatment Effect (ATE) is conclusive and positive in the metrics of interest, such as conversions or profit, the promotion is rolled out for everyone. However, many promotion campaigns are only profitable for a small fraction of the population. In these cases, marketers can still run a profitable campaign, or at the very least, break even, if they only target the subgroups that show favorable unit economics, i.e., the trade-off between extra conversions and profit. For that, one requires good estimates of the effect of a promotion on population subgroups. 

Uplift modeling~\cite{radcliffe1999differential, devriendt2018literature} allows estimating the treatment effect of giving or not a promotion at the subgroup level, i.e., the Conditional Average Treatment Effect (CATE), from the data. Furthermore, this allows personalizing promotions to individuals that get the highest benefit in terms of an outcome of interest~\cite{olaya2020survey}. In recent years, uplift modeling has seen uptake by large e-commerce platforms such as Uber~\cite{zhao2019uplift}, Spotify~\cite{fernandez2022comparison}, Amazon~\cite{mondal2022aspire}, Criteo~\cite{diemert2018large}, and Booking.com~\cite{goldenberg2020free}. 

However, most of the \textbf{related work} has focused on maximizing conversion rate, which assumes that the costs per promotion are constant, which is rarely the case~\cite{haupt2022targeting}. Moreover, maximizing conversions can be even counterproductive to a promotion's business objectives as it may negatively impact the campaign's profitability. 
For example, if a promotion results in higher costs that outweigh the revenue generated from additional conversions, it can lead to a negative profit.
Thus, for uplift modeling to aid e-commerce, it requires explicitly measuring the subgroups' profit or the trade-off between conversions and costs.

This issue has been taking more attention in recent years~\cite{haupt2022targeting}, with more works focusing on estimating business metrics~\cite{gubela2017revenue,zhao2019uplift,gubela2020response,haupt2022targeting,verbeke2023or,lessmann2021targeting}, e.g., profit uplift. Nonetheless, these approaches suffer from several technical limitations: 1) they usually require using more than one machine learning model, all optimized independently; 2) profit values are usually zero-inflated (too many zeros), which most standard machine learning algorithms are not prepared to deal with~\cite{haupt2022targeting}---this arises because non-conversions have virtually no cost for several promotion types, such as discounts or coupons; 3) search data in e-commerce platforms are in general noisy, due to low conversion values (circa 3\%), scrapes and crawlers~\cite{goldenberg2020free}, and many-to-one mapping of searches, i.e., a purchase can have many associated searches. 

Other cost-aware approaches optimize the whole campaign to maximize volume, e.g., conversions, with financial constraints, e.g., limited budget~\cite{goldenberg2020free,albert2022commerce,ai2022lbcf,yan2023end}. However, these approaches face similar problems as above while adding an extra layer of complexity to the problem of promotion allocation. The only work we know that considers tackling the search data problem is \citet{goldenberg2020free}, which proposes criteria based on a ratio of the effect in conversion and loss (negative profit) while only using converted data.

In \textbf{this work}, we focus on the specific scenario of profit estimation when the promotion costs are zero if the item is not purchased, i.e., response-dependent costs \cite{haupt2022targeting}. We propose Incremental Profit per Conversion (IPC), a measure of the efficiency of a promotion as an investment for a subgroup in terms of unit economics, i.e., it quantifies the trade-off between conversions and their profit, contrary to most related work, which focuses on the expected total profit per subgroup at a search or visit level. Moreover, IPC and its estimator aim to address three main limitations of the related literature mentioned above.

The \textbf{main contributions} of this work can be summarized as follows:
\begin{enumerate}
    \item \textbf{Proposing the Incremental Profit per Conversion (IPC)} as a measure of personalized promotion success in cases where promotion campaigns incur costs solely associated with conversions.
    \item \textbf{Defining a response transformation that allows estimating IPC using only converted data and a single (agnostic) machine learning model}. These properties help to mitigate the zero-inflated profit values, many-to-one mapping between search and purchase data, noise inherent in search data while using only one agnostic machine learning model, and around 100 faster training times.
\end{enumerate}

\section{Uplift modeling background}

Uplift modeling is a technique that analyzes how a treatment affects the outcome of interest of an individual---such as conversion or profit in e-commerce. It allows for the identification of individuals who exhibit a greater response sensitivity to the treatment. Data for uplift modeling is typically collected through Randomized Controlled Trials (RCT) such as A/B tests---where individuals are randomly split between treatment and control groups---or from previous such experiments. 

Uplift modeling can be seen as a form of causal inference, and as such, we present it within the potential outcomes framework of the Neyman-Rubin causal model~\cite{rubin1974estimating}. In the case of two possible treatments---$T=0$ representing the control group, and $T=1$ the treatment group---we can define an potential outcome $Y$ for each $i$ as $Y_i(t)$ for each treatment group and obtain the Individual Treatment Effect (ITE) as:
\begin{equation}
    Y_i(T=1) - Y_i(T=0).
\end{equation}
Estimating the ITE is only feasible if we had access to the outcomes of both treatment and control for the same individual, which is impossible in practice---also known as the fundamental problem of causal inference \cite{holland1986statistics}. However, the Conditional Average Treatment Effect (CATE) is a related quantity defined as the expectation of the difference of the potential outcomes for an individual's characteristics $\mathbf{X}=\mathbf{x}$:
\begin{equation}\label{eq:cate}
    \tau_{Y}(x) = \mathbb{E}[Y(1)-Y(0) \mid \mathbf{x}],
\end{equation}
which, under the potential outcomes assumptions, can be computed as:
\begin{equation*}
    \tau_{Y}(x) = \mathbb{E}[Y \mid \textbf{x}, T=1] - \mathbb{E}[Y \mid \textbf{x}, T=0].
\end{equation*}

\subsection{Conversion and profit uplift modeling}
Throughout the remainder of this work, we consider a dataset of the form: $D = ( \textbf{X}, T, C, \Pi ) = \{(\textbf{x}_i, t_i, c_i, \pi_i)\}_{i=1}^{n}$ of $n$ \emph{i.i.d.} treatment units. $\textbf{X}$ represents the vector of the context of a treatment unit, which in the case of e-commerce is made of the individual's and item's characteristics, $T$ is the assigned treatment---also called promotion when we refer to the e-commerce setting---and $C \in \{0, 1\}$ and $\Pi \in \mathbb{R}$ are two outcomes of interest, the conversion, and profit, respectively. \\

Profit is defined as $\pi_i = revenue_i - cost_i$, with the $cost$ representing the $cost$ associated with treatment and context. Treatment costs can generally be fixed per treatment or response-dependent \cite{haupt2022targeting}. This work considers the scenario where only a conversion has associated costs (only response-dependent costs), i.e., $C=0 \Rightarrow \Pi = 0$. This setup is common in e-commerce settings, where treatments are in the form of promotions with coupons or discounts, only incurring costs in case of a positive conversion, and fixed costs per treatment are negligible. Also, we refer to the converted-only dataset as $D^{C=1}= \{ (\textbf{x}_i, t_i, c_i, \pi_i) \in D \mid C=1 \}$. 

Building upon the previous definitions, we can define the CATE of profit according to  \eqref{eq:cate}, as:
\begin{equation}
    \tau_{\Pi} = \mathbb{E}[\Pi \mid \textbf{x}, T=1] - \mathbb{E}[\Pi \mid \textbf{x}, T=0],
\end{equation}
which represents the incremental profit obtained by treating a subgroup with characteristics $\textbf{X}=\textbf{x}$ with treatment $T=1$. The intuition behind the CATE of profit is that $\mathbb{E}[\Pi \mid \textbf{x}, T=0]$ represents the baseline profit (without promotion), while $\mathbb{E}[\Pi \mid \textbf{x}, T=1]$ represents the profit when the promotion is given. The difference between these two signifies the extra profit the e-commerce would earn if it were to implement the promotion for context $\textbf{x}$.  

\section{Incremental Profit per Conversion (IPC)}

In this section, we propose Incremental Profit per Conversion (IPC)---\textbf{CATE of profit per conversion}---a new measure to estimate the incremental profit of a promotion for a context $\textbf{X}=\textbf{x}$, in the case where \textit{non-converted individuals carry no cost}. Common examples of cost scenarios in e-commerce promotions include coupons, discounts, and emailed offers~\cite{sahni2017targeted}.

Moreover, we show in Section~\ref{sec:ipc} that estimation of IPC only requires converted data (or purchased transactions) $D^{C=1}$. This estimation helps mitigate common issues with e-commerce data, such as, the noise associated with non-converted data like crawlers and scrappers \cite{goldenberg2020free}, the zero-inflated profit or revenue data, and the many-to-one problem of mapping searches to purchases. The underlying intuition is that non-conversions should not significantly impact the profitability per conversion of a promotion.

Formally, we can define IPC as:
\begin{equation}\label{eq:iprofit_per_conversion}
\begin{split}
IPC(\textbf{x}) &= \frac{\tau_{\Pi}(\textbf{x})}{Pr(C=1 \mid \textbf{x})}  \\
& =\frac{\mathbb{E}[\Pi \mid \textbf{x}, T=t]-\mathbb{E}[\Pi \mid x, T=0]}{Pr(C=1 \mid \textbf{x})},
\end{split}
\end{equation}
where $Pr(C=1 \mid \textbf{x}) = Pr(C=1 \mid \textbf{x}, T=0 \vee T=1)$ is the probability of a conversion occurring in control or treatment groups. The intuition is that the highest incrementality comes from searches with the highest profit per conversion. 
This metric quantifies the effectiveness of a promotion targeting strategy and can be combined with a decision rule that sets spending goals per conversion or ranks contexts (which is a common approach in uplift modeling).

Next, we show how a feature transformation allows us to estimate IPC per context $\textbf{X}=\textbf{x}$ using only converted data $D^{C=1}$.

\subsection{Incremental profit per conversion response transformation}\label{sec:ipc}

To estimate the incremental profit per conversion, we propose a response transformation that can only be applied when only positive conversions carry a cost, i.e., $C_i=0 \Rightarrow cost_i = 0$. Moreover, this assumption ensures that the \textbf{transformation only relies on converted data}, and the transformed response variable $Z$ can be defined per instance as follows:

\begin{equation}\label{eq:response_transform}
z_i = \left\{\begin{matrix}
-\frac{1}{\Pr(T=0 \mid \textbf{x})}\pi_i & \text{if } T_i =0 \:\wedge \: C_i=1 \\ 
+\frac{1}{\Pr(T=1 \mid \textbf{x})}\pi_i & \text{if } T_i =1 \:\wedge \: C_i=1 \\ 
\text{remove instance} & \text{if } C_i=0
\end{matrix}\right.
\end{equation}
where $\pi_i=r_i -cost_i$ is the profit associated with instance $i$, and where the third line ($C_i=0$) represents that only converted samples are necessary for this approach. $\Pr(T=0 \mid \textbf{x})$ and $\Pr(T=1 \mid \textbf{x})$ are the propensities associated with treatment and control samples, which in the case of a randomized trial of equal-sized groups do not depend on $\vec{x}$ and correspond to $\Pr(T=0) = \Pr(T=1) = 0.5$. 

In Appendix~\ref{appendix:proof} we show that the expected value of $\mathbb{E}[Z \mid \textbf{x}, C=1] = \mathbb{E}[IPC(\textbf{x})]$. To estimate IPC, a standard supervised learning regression algorithm can be employed with the target variable set as $Z$, thereby providing a valid estimate of Eq.~\eqref{eq:iprofit_per_conversion}.

This outcome transformation shares similarities with the one proposed by~\citet{gubela2020response}; however, their approach considers profit maximization over the entire data $D$, whereas we restrict the data to converted data only  $D^{C=1}$. We argue and show over an example dataset in Section~\ref{sec:example} that if the objective is to make the best investments in terms of unit economics, IPC and our response transformation are better measure of investment efficiency.

\subsection{Example computation and other measures}\label{sec:example}

In Table~\ref{tab:example_data}, we present an example dataset that showcases the estimation of IPC and illustrates why it represents a measure of the efficiency of a promotion campaign. 

From this example dataset, two observations become apparent. Firstly, the lower-profit purchases in the treatment group (instances $5$ and $6$) compensate for one purchase with higher profits in the control group (instance $3$). Secondly, non-converted users should not influence the profit per conversion, as their profit is zero.

\begin{table}																
\caption{Promotion campaign example A/B experiment dataset. The description of the columns is \{instance index; treatment; context; conversion; profit; response transform; propensity; probability of converting; probability of converting given treatment \}. This dataset contains $6$ instances, two treatments ($0$ and $1$) with equal propensity ($\Pr(t \mid x)=0.5$), and one context feature $x=1$.}																
\label{tab:example_data}																
\begin{tabular}{@{}lrrrrrrrr@{}}																
\toprule																
 $i$ & $t$	&	$x$	&	$c$	&	$\pi$\footnotesize	 (\euro)	&	$z$	&	 $\Pr(t | x)$	&	$\Pr(C=1 | x)$	&	$\Pr(C=1 | x, t)$	\\ \midrule	
1 & 0	&	1	&	0	&	0	&	-	&	0.5	&	0.5	&	0.33	\\	
2 & 0	&	1	&	0	&	0	&	-	&	0.5	&	0.5	&	0.33	\\	
3 & 0	&	1	&	1	&	10	&	-20	&	0.5	&	0.5	&	0.33	\\ \hline	
4 & 1	&	1	&	0	&	0	&	-	&	0.5	&	0.5	&	0.66	\\	
5 & 1	&	1	&	1	&	8	&	16	&	0.5	&	0.5	&	0.66	\\	
6 & 1	&	1	&	1	&	8	&	16	&	0.5	&	0.5	&	0.66	\\ \bottomrule	
\end{tabular}																
\end{table}																			

We now show that IPC captures this trend, contrary to other proposed measures.

In the this example, for context $X=1$ the IPC value can be estimated by \eqref{eq:iprofit_per_conversion} or \eqref{eq:response_transform}:
\begin{equation*}
IPC(X=1) = \frac{32-20}{3} = 4 \text{ \euro}/conversion,
\end{equation*}

which means that for each conversion, we earn $4$ extra \euro{} if the promotion $T=1$ would be applied. IPC assumes that the control group of $T=0$ represents a purchase generated without a promotion and the treatment $T=0$ the two purchases obtained with a promotion. On average, the profit would increase if the context $X=1$ would be treated with $T=1$. \\

To gain a deeper understanding of IPC, we compare it with other related measures. The most related measure is the \emph{CATE of profit} as proposed by \citet{gubela2020response,haupt2022targeting} which gives:
\begin{equation*}
\tau_\Pi (X=1) = \frac{32}{3} - \frac{20}{3} = 2 \text{ \euro},
\end{equation*}
which is an estimate of expected profit per instance. Hence, CATE of profit values contexts with many conversions. To illustrate this point clearly, imagine that Table~\ref{tab:example_data} would have another context $X=2$ that has the same number of converted purchases as $X=1$, i.e., one in control and two in treatment, and $100$ more non-converted users. In that case $IPC(X=2,T=1) = 4 \text{ \euro}/conversion$ would remain the same, while $CATE_\Pi (X=2, T=1) = 0.06 \text{ \euro}$. Therefore, CATE profit would value more $X=1$, even though $X=1$ and $X=2$ have the same profit per conversion.



In the case of the \emph{retrospective estimator}~\cite{goldenberg2020free}, the maximization objective is the following:
\begin{equation*}
    \frac{\tau_{C}(X=1)}{-\tau_{\Pi}(X=1)} = \frac{2 S(1) -1}{ [1- S(1)]\bar{\pi}_{0} -S(1)\bar{\pi}_{1} } = \frac{1}{-6} = -0.17 \text{ \euro}^{-1},
\end{equation*}
where $S(1)= \Pr(T=1 \mid X=1, C=1)$ and $\bar{\pi}_{t} = \mathbb{E}[\Pi | C=1, T=t]$ is the average profit for all units treated with $T=t$. This implies that one acquires $-0.17$ extra conversions per extra euro spent if $X=1$ is treated, i.e., $5.9$ \text{ \euro} are earned per extra conversion. Note that the simplified version of the retrospective estimator assumes constant $\bar{\pi}_{0}$ and $\bar{\pi}_{1}$, which makes the sorting of contexts only dependent on the sorting given by $S(1)$.

Finally, the last work that has similarities with IPC is of~\citet{zhao2022multiple}, where it optimizes for:
\begin{equation*}
\tau(X=1,\pi) = \mathbb{E}[\pi_1 Y_1 - \pi_0 Y_0 \mid X=1]  =  \tau_{\Pi}(X=1),
\end{equation*}
when the impression costs are zero. In this case, this approach leads to an equivalent estimate as of~\citet{gubela2020response}; however,~\citet{zhao2022multiple} approach assumes that the value and cost per conversion are fixed and not random variables that can be different for the same context.
\section{Experiments}

To evaluate the accuracy of the IPC estimator, we compare its performance against several State-Of-The-Art (SOTA) uplift and causal inference methods in terms of their ability to rank instances based on our outcome of interest, i.e., profit. The quality of these methods is measured using the Qini coefficient and the area under the Qini curve~\cite{radcliffe2007using}.

To conduct the evaluation, we generate a synthetic dataset that simulates a discount coupon campaign for an e-commerce platform. 

This section is organized as follows: first, we present the SOTA methods that were used for comparison. Next, in Section~\ref{sec:synthetic}, we outline the process of generating the synthetic data. Finally, in Section~\ref{sec:results}, we present the results of the IPC and SOTA methods applied to the synthetic data, using the Qini coefficient and runtime values as evaluation metrics.\\

The causal inference and uplift modeling \textbf{methods} used to compare against include: 
\begin{itemize}
    \item \emph{Retrospective estimator} \cite{goldenberg2020free, teinemaa2021upliftml} - this method proposes a sorting CATE of conversion per CATE of loss (negative profit). This method, similar to IPC, also only requires converted data. The standard implementation of this estimator assumes that the revenue in the control and treatment groups is similar for the same context $\textbf{x}$. It can utilize an agnostic machine-learning model for estimation.
    \item \emph{Response transformations} RDT and CRVTW \cite{gubela2020response} - \citet{gubela2020response} proposed several profit estimation techniques, among which Revenue Discretization Transformation (RDT) and Continuous Response Variable Transformation with Weightings (CRVTW) showed promising results. RDT transforms the problem into a binary classification task, where the objective is to predict whether the profit is above zero or not. CRVTW is a response transformation similar to IPC but applied to search data.
    \item \emph{Meta-learners} - this comprises the S-, T-, X-, R-, and Doubly Robust-learners \cite{hansotia2002incremental, kunzel2019metalearners, nie2021quasi, kennedy2020towards}. These methods use agnostic machine learning models to estimate the CATE for binary or continuous outcomes. We used the implementation provided by the causalML package \cite{chen2020causalml}. 
    \item \emph{Causal Forest} (CF) \cite{wager2018estimation} - CF is an adaptation of the standard random forest algorithm that incorporates a procedure for estimating causal effects while maintaining several desirable statistical properties. We utilized the EconML implementation \cite{econml} for this method.
\end{itemize}

The machine learning algorithm used for all these methods was the Gradient Boosting Regressor from \texttt{scikit-learn} \cite{scikit-learn} configured with $1000$ iterations and early stopping.

\subsection{Synthetic coupon campaign generation}\label{sec:synthetic}
The objective of generating synthetic data is to simulate an e-commerce A/B test---Randomized Controlled Trial---conversion dataset for a percentage discount coupon campaign, where profit and cost are zero for non-conversions, i.e., $C_i=0 \rightarrow \Pi_i=0, cost_i =0$. This represents a common setting where users receive percentage coupon for any item they want.

To generate the synthetic dataset, we follow a three-step process: first, we generate the conversion dataset; second, we generate the revenue based on the conversion features; and finally, we compute the profit by applying a percentage discount to the treated units.

The \emph{conversion dataset} is characterized by: the number of samples, its conversion rate in the control group, and the relationship between the features and conversion. The features can be divided into three types:
\begin{enumerate}
    \item \emph{Positive treatment effect features} - features that positively affect the conversion.
    \item \emph{Informative features} - features that influence the conversion rate but do not distinguish between treatments.
    \item \emph{Uninformative features} - features that do not affect the conversion rate or the treatment effect.
\end{enumerate}
The \emph{revenue generation} calculates the revenue using a subset $S_R$ of the two first types of conversion features through the following function: 
\begin{equation*}
R(\textbf{x}_i) = \left\{\begin{matrix}
\exp(\sum_{x_ji \in \textbf{x}_i} \mathbbm{1}(X_j \in S_R) \cdot x_{ji} + \epsilon & \text{if } C_i=1 \\ 
0 & \text{if } C_i=0  
\end{matrix}\right.
\end{equation*}
where $ \mathbbm{1}(X_j \in S_R)$ is an indicator function saying that the feature $X_j$ is in the subset of revenue features $S_R$, $x_{ji}$ is the value of that feature for instance $i$, and $\epsilon$ is normally distributed noise term. The exponential function represents the distribution of revenues, which usually follows an exponentially decaying distribution. 

Finally, the $cost$ of the coupon is added as a percentage $d$, to compute the profit for each conversion, with the following form:
\begin{equation*}
\Pi(\textbf{x}_i) = \left\{\begin{matrix}
R(\textbf{x}_i) & \text{if } T_i=0 \\ 
R(\textbf{x}_i)*(1-d) & \text{if } T_i=1 \\
\end{matrix}\right.
\end{equation*}

The \textbf{data for this experiments} is generated with $200\,000$ samples; one control and one treatment group; a propensity of $0.5$ per treatment; a control group conversion rate of $3\%$; 13 features---$3$ with Positive Treatment effect, $5$ informative, $5$ irrelevant; only two features in the revenue set (one informative and one with positive treatment effect); a noise term with 10\% lower standard deviation than those two features; and a $10\%$ discount.

\subsection{Results and discussion}\label{sec:results}
\begin{figure}[h]
  \centering
  \includegraphics[width=\linewidth]{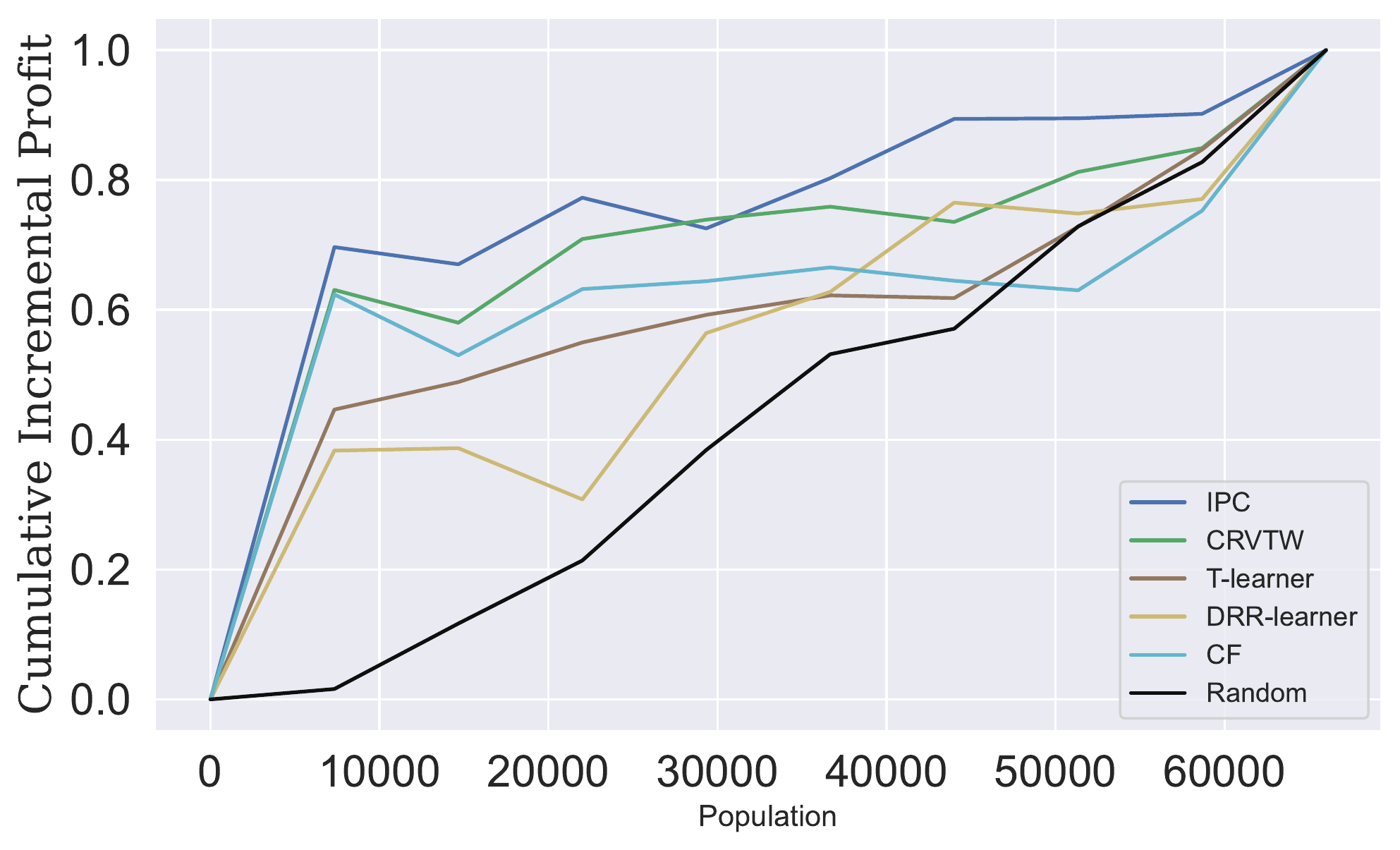}
  \caption{Normalized Profit Qini curves of the top-5 models for a 70\%-30\% train-test split on the synthetic discount coupon campaign data.}
  \label{fig:qini_curve}
\end{figure}

\begin{figure}[h]
  \centering
  \includegraphics[width=\linewidth]{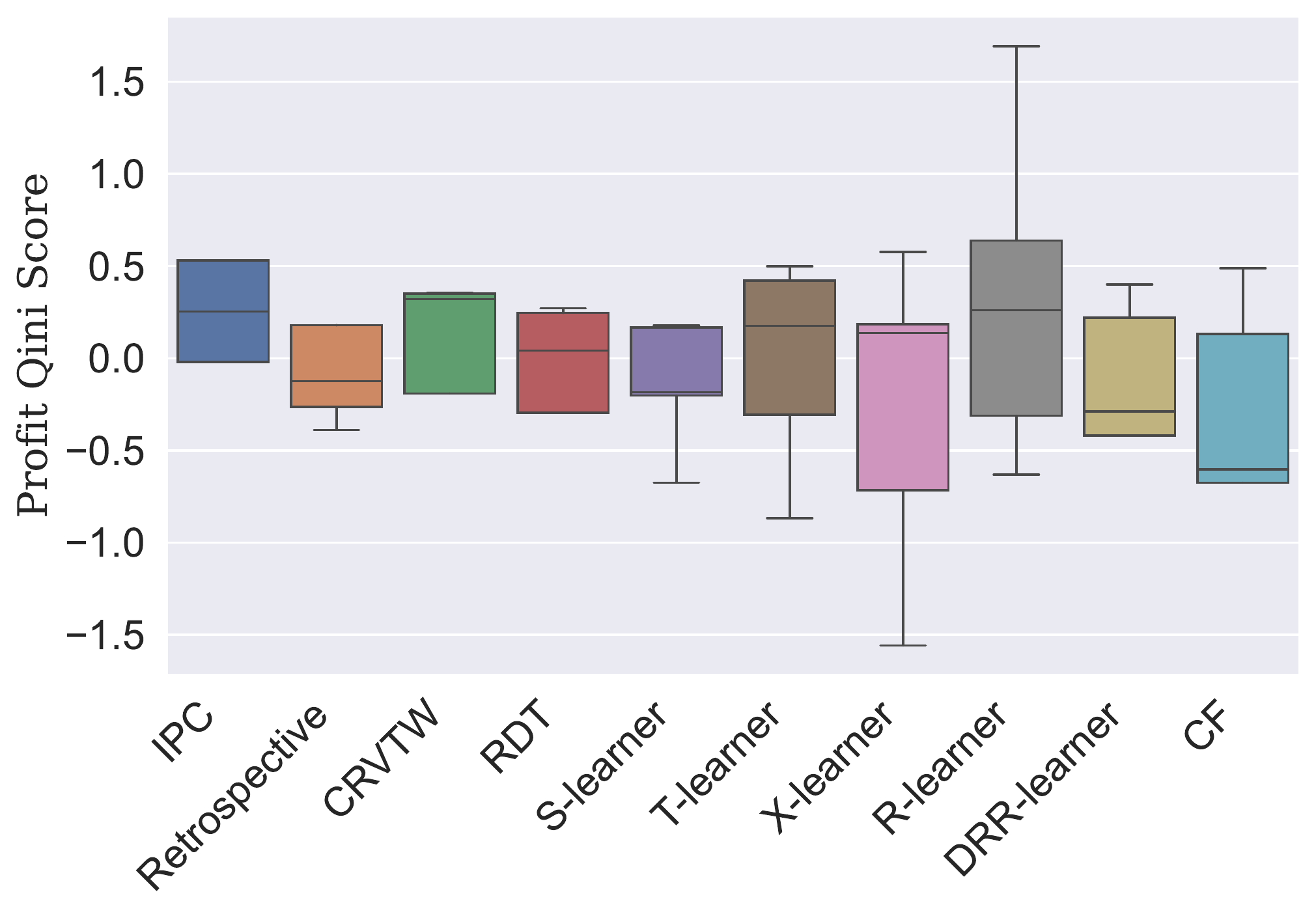}
  \caption{Box plot of the Profit Qini scores for a 5-fold crossvalidation on the synthetic discount coupon campaign data. }
  \label{fig:violin_qini_score}
\end{figure}

\begin{figure}[h]
  \centering
  \includegraphics[width=\linewidth]{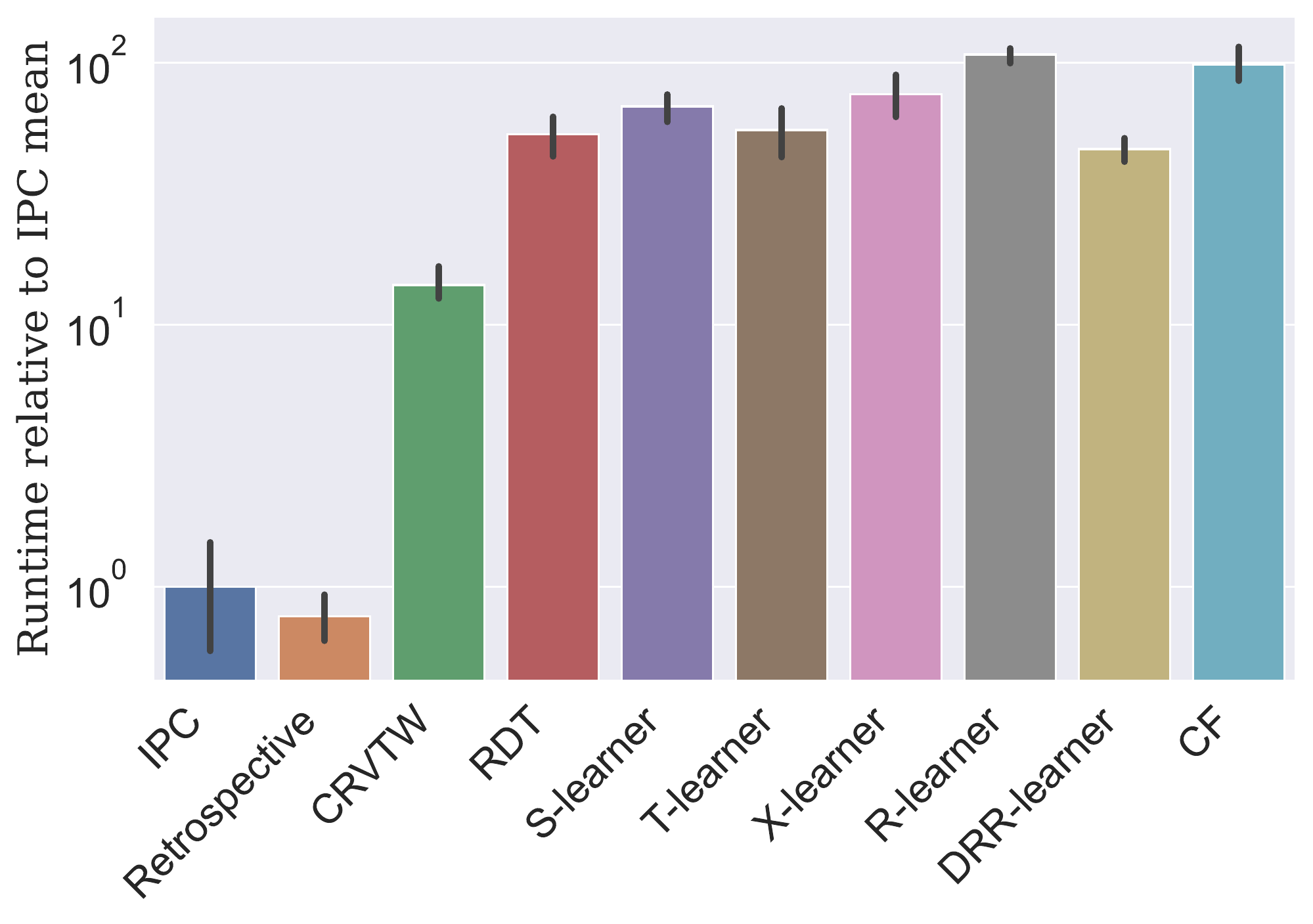}
  \caption{Runtime relative to IPC mean for all algorithms from 5-fold on synthetic discount coupon campaign data.}
  \label{fig:bar_runtime}
\end{figure}

In this analysis, we compare the sorting of instances by profit and the overall runtime for all algorithms over the synthetically generated data.
To this end, the Qini score of the profit and runtime results for $5$-fold cross-validation is presented in Figure~\ref{fig:violin_qini_score} and \ref{fig:bar_runtime}, respectively, while a Qini curve for the top-5 algorithms over a $70\%$-$30\%$ train-test split is in Figure~\ref{fig:qini_curve}.
Regarding the model performance in terms of Qini-scores, our first observation is the absence of a clear winning algorithm for this problem. In fact, some models even exhibit Qini scores below random sorting, indicating subpar performance. However, Figure~\ref{fig:violin_qini_score} shows that IPC is in the top range of performances with lower variance than other methods. This is also noticeable in Figure~\ref{fig:qini_curve}, where IPC seems to be more efficient at targeting smaller percentages such as 10\% and 20\% of the population. The R-learner has a similar mean performance, although very inconsistently over the folds. Finally, it is evident that the retrospective assumption of similar revenue in the control and treatment groups does not hold for this dataset. This is clearly demonstrated by the explicitly generated revenue, which incorporates a positive treatment effect feature and an informative feature. Consequently, the models based on this assumption exhibit poor performance.

Regarding the runtime of Figure~\ref{fig:bar_runtime}, it can be seen that IPC and retrospective have the lowest runtimes. This was expected, as both models only use converted data to train. 

Altogether, one can see that the IPC estimator can perform on par or better than other SOTA algorithms, while having a runtime that stands out by being two orders of magnitude lower than competitors. 

\section{Conclusions}
In this paper, we introduced Incremental Profit per Conversion (IPC), a measure for uplift modeling that allows ranking promotion campaigns' profitability when the cost of a non-converted promotion is zero. Moreover, we proposed a response transformation that is a valid estimator of IPC and only requires converted data to compute if the propensities are known---a common scenario when uplift models use A/B experiments data to train. 

IPC presents two main advantages. First, while most uplift literature deals with conversion uplift, the focus on profit gets closer to the business metrics of interest, such as keeping profitability and costs of promotion campaigns under control. Second, using only converted data is an advantage for e-commerce datasets, where conversion rates were around $2.5\%$~\cite{chevalier23} in 2022. This last property allows dealing with the data imbalance directly by using only converted data, removing the noise and biases present with non-converted data, such as bots, and faster training times for machine learning models.

To evaluate the quality of our proposed measure and estimate, we compare our approach with several SOTA algorithms on a synthetic dataset that simulates discount coupon campaigns. The results show that IPC performs on par or better than other algorithms, taking 100 times less time. In future work, we aim to extend our analysis to real-world datasets on promotion campaigns.

\section{Acknowledgments}

We would like to thank Dima Goldenberg, Igor Spivak, Itsik Adiv, and Carlos Herrero for the insightful feedback and comments, Ulf Schnabel and Ioannis Kangas for their guidance, and Erdem Kulunk for his support throughout this project.
\bibliographystyle{ACM-Reference-Format}

\appendix

\section{Appendix}

\subsection{Derivation of IPC from response transformation}\label{appendix:proof}

We show how the expected value of response transformation of \eqref{eq:response_transform}, is the same as the Incremental Profit per Conversion (IPC) of \eqref{eq:iprofit_per_conversion}. First we need to define the set of possible values of $z$ and $\pi$. $\mathcal{Z}^{T=t}$ represents all possible values of $z$ present in $D^{C=1}$ given the treatment $t$, and $\mathcal{P}$ represents all the possible profit values present in the dataset $D^{C=1}$, the proof is a follows:
\begin{equation}
\begin{split}
&E[Z \mid \textbf{x}, C=1] = \\
&= \sum_{\pi \in \mathcal{P}} \frac{\pi \Pr(\Pi=\pi, T=1 |\textbf{x}, C=1)}{\Pr(T=1 \mid \textbf{x})} - \frac{\pi \Pr(\Pi=\pi, T=0 |\textbf{x}, C=1)}{\Pr(T=0 \mid \textbf{x})} \\
& = \sum_{\pi \in \mathcal{P}} \frac{\pi \Pr(\Pi=\pi \mid \textbf{x}, C=1, T=1 ) \Pr(T=1 \mid \textbf{x}, C=1 )}{\Pr(T=1 \mid \textbf{x})}\\
& - \sum_{\pi \in \mathcal{P}} \frac{\pi \Pr(\Pi=\pi \mid \textbf{x}, C=1, T=0 ) \Pr(T=0 \mid \textbf{x}, C=1 )}{\Pr(T=0 \mid \textbf{x})}\\
& = \sum_{\pi \in \mathcal{P}} \frac{\pi \Pr(\Pi=\pi \mid \textbf{x}, C=1, T=1 ) \Pr(C=1 \mid  \textbf{x}, T=1 )}{\Pr(C=1 \mid \textbf{x})}\\
& -\sum_{\pi \in \mathcal{P}} \frac{\pi \Pr(\Pi=\pi \mid \textbf{x}, C=1, T=0 ) \Pr(C=1 \mid  \textbf{x}, T=0 )}{\Pr(C=1 \mid \textbf{x})}\\
& =  \left [ \sum_{\pi \in \mathcal{P}} \frac{\pi \Pr(\Pi=\pi \mid \textbf{x}, C=1, T=1 ) \Pr(C=1 \mid  \textbf{x}, T=1 )}{\Pr(C=1 \mid \textbf{x})} \right.\\
& + \left. \sum_{\pi \in \mathcal{P}} \frac{\pi \Pr(\Pi=\pi \mid \textbf{x}, C=0, T=1 ) \Pr(C=0 \mid  \textbf{x}, T=1 )}{\Pr(C=1 \mid \textbf{x})} \right] \\
& -  \left [ \sum_{\pi \in \mathcal{P}} \frac{\pi \Pr(\Pi=\pi \mid \textbf{x}, C=1, T=0 ) \Pr(C=1 \mid  \textbf{x}, T=0 )}{\Pr(C=1 \mid \textbf{x})} \right.\\
& + \left. \sum_{\pi \in \mathcal{P}} \frac{\pi \Pr(\Pi=\pi \mid \textbf{x}, C=0, T=0 ) \Pr(C=0 \mid  \textbf{x}, T=0 )}{\Pr(C=1 \mid \textbf{x})} \right] \\
&= \frac{\mathbb{E}[\Pi \mid \textbf{x}, T=1] - E[\Pi \mid \textbf{x}, T=0]}{\Pr(C=1 \mid \textbf{x})} \\
&= IPC(\textbf{x})
\end{split}
\end{equation}
The first expression is given to us by the definition of the response transformation and law of iterated expectations. Then, the second expression is obtained by applying the chain rule. After that, the third expression comes from using the Bayes rule on $\Pr(T=1 \mid x, C=1 )$, similarly to \cite{goldenberg2020free}. Finally, the fourth expression appears by considering that the profit is always zero for all non-conversions.

\end{document}